\documentclass[11pt,a4paper]{article}

\usepackage[utf8]{inputenc}
\usepackage[T1]{fontenc}
\usepackage{lmodern}
\usepackage[english]{babel}
\usepackage{lettrine}
\usepackage{enumitem}

\usepackage{amsmath,amssymb,amsthm}
\usepackage{mathtools}

\usepackage{graphicx}
\usepackage{float}
\usepackage{subfig}

\usepackage{booktabs}
\usepackage{multirow}
\usepackage{array}

\usepackage{algorithm}
\usepackage{algorithmic}

\usepackage[numbers,sort&compress]{natbib}
\usepackage[colorlinks=true,linkcolor=blue,citecolor=blue,urlcolor=blue]{hyperref}

\usepackage[margin=1in]{geometry}

\usepackage{setspace}

\usepackage{titling}

\theoremstyle{definition}

\theoremstyle{remark}

\providecommand{\keywords}[1]{\textbf{Keywords:} #1}

\title{From Projection to Prediction: Beyond Logits for Scalable Language Models}

\author{
    Jianbing Dong\thanks{Corresponding author}\\
    \textit{NVIDIA}\\
    jianbingd@nvidia.com
    \and
    Jianbin Chang\\
    \textit{NVIDIA}\\
    jianbinc@nvidia.com
}



\date{}

\begin{document}

\pretitle{
  \begin{center}
  \rule{\textwidth}{4pt}
  \\[0.5em]
  \LARGE\bfseries
}
\posttitle{
  \rule{\textwidth}{1.5pt}
  \end{center}
  \vspace{1em}
}
\maketitle

\begin{abstract}
Training Large Language Models (LLMs) typically involves a two-stage pipeline at the output layer: 
hidden states are projected into vocabulary logits via a linear transformation (lm\_head), followed 
by cross-entropy loss computation against target tokens. While conceptually simple, this design incurs
substantial overhead. The intermediate logits tensor, with dimensions proportional to batch size, sequence length,
and vocabulary size, must be fully materialized in GPU memory, even though only one target token per position
is ultimately used. This leads to significant memory footprint and bandwidth comsumption, limiting scalability and
slowing training throughput.

In this work, we introduce a novel approach to integrates the output projection and loss prediction into a single
operation. By directly computing the loss from hidden states and target tokens, our approach bypasses explicit logits
materialization. This design reduces memory usage and alleviates bandwidth pressure. Experiments
on LLM training demonstrate that our method achieves substantial memory savings and measurable speedups compared to 
the standard two-stage pipeline, enabling large batch sizes and longer sequences without sacrificing accuracy. Our work 
highlights the benefits of rethinking the boundary between projection and prediction, offering a practical systems 
optimization for efficient LLM training.
\end{abstract}


\section{Introduction}
\label{sec:introduction}
\lettrine{L}{arge} language models (LLMs) have rapidly advanced the state of natural language processing, 
powering applications in dialogue systems \cite{yi2025surveyrecentadvancesllmbased,legashev2025}, 
code generation \cite{stefanobistarelli}, and scientific discovery \cite{yanbozhangexploringtherole}. 
Their success has been enabled by scaling both model size and training data, but this scaling comes at a steep computational 
and memory cost \cite{ICLR2024Thecostof,JulianBuchelEfficientScaling,ahmad2025optimizing}. 
As models grow to billions of parameters and vocabularies reach hundreds of thousands of 
tokens, even seemingly routine components of the training pipeline can become critical bottlenecks \cite{ChaofanTaoScalingLaws}.

A particularly important bottleneck lies in the output layer. In every standard LLM training workflow, 
hidden states are projected into vocabulary logits through a large linear transformation known as the 
\texttt{lm\_head}. These logits are then consumed by a cross-entropy loss function to predict target 
tokens. This two-stage design is not an implementation detail but a canonical requirement: 
the \texttt{lm\_head} projection and cross-entropy loss together define how probability distributions 
over vocabularies are learned \cite{10888877}. However, prevailing implementations invariably insist on materializing 
the entire logits tensor in GPU memory \cite{IJCAI24TaehoKim}, a practice that inflates computational overhead and saturates 
bandwidth despite its limited utility. The tensor's dimensions scale with batch size, sequence length, and 
vocabulary size, leading to substantial memory footprint and bandwidth consumption. 
Crucially, only one target logit per position is ultimately used in the numerator of the loss, 
meaning the vast majority of the computed logits are discarded. This mismatch between computation 
and utility limits scalability and slows training throughput.

Recent research has explored kernel fusion and operator optimization in other parts of the LLM training 
stack, such as online softmax \cite{onlinesoftmax}, attention mechanisms \cite{dao2022flashattention, dao2023flashattention2}, 
and optimizer updates \cite{rajbhandari2020zeromemoryoptimizationstraining,rajbhandari2021zeroinfinitybreakinggpumemory}. 
These efforts highlight the importance of reducing redundant memory movement and improving arithmetic intensity. 
To the best of our knowledge, the output projection and loss computation — despite being a mandatory stage in every 
LLM pipeline — have remained largely untouched \cite{10888877}. Existing frameworks continue to treat 
the \texttt{lm\_head} and cross-entropy loss as separate modules, leaving untapped opportunities for optimization.

In this work, we propose a fused kernel implementation that integrates the output projection and loss 
prediction into a single operation. By directly computing the loss from hidden states and target tokens, 
our approach bypasses explicit logits materialization while preserving the standard training semantics. 
This design reduces memory usage, alleviates bandwidth pressure, and improves throughput. Experiments 
on LLM training demonstrate that our method achieves substantial memory savings and measurable speedups 
compared to the canonical two-stage pipeline, enabling larger batch sizes and longer sequences without 
sacrificing accuracy.

Our contributions underscore the value of rethinking the boundary between projection and prediction. 
By collapsing these stages into a unified kernel, we provide a practical systems optimization that 
addresses a critical bottleneck in the universally adopted LLM training workflow. Beyond immediate 
performance gains, this work contributes to the broader effort of making large-scale language modeling 
more efficient, sustainable, and accessible.

\section{Related Work}
\label{sec:related_work}
Optimizing the efficiency of large language model training has been a central focus of recent systems research. 
A recurring theme is the reduction of redundant memory movement and the fusion of operations to improve arithmetic 
intensity \cite{11161662}. These efforts demonstrate that even small changes to core primitives can yield 
substantial performance gains at scale.

\subsection{Softmax Optimizations}
Online Softmax \cite{onlinesoftmax} introduced a memory-efficient reformulation of the softmax operator that reduces 
redundant memory accesses by computing the normalization term in a single pass over the input vector. This approach 
improves GPU performance for bandwidth-bound workloads, achieving up to 1.3$\times$ speedup compared to the standard 
safe softmax implementation. Furthermore, by fusing Online Softmax with the commonly used Top-K operation in beam 
search, they demonstrated up to 5$\times$ performance gains. This work highlights how kernel-level optimizations in 
seemingly routine components of the training stack can deliver substantial efficiency improvements, influencing 
subsequent research on operator fusion and memory-aware kernel design in LLM systems.

\subsection{Attention Mechanism Optimizations}
The attention mechanism has been another major target for kernel fusion. FlashAttention \cite{dao2022flashattention, dao2023flashattention2} 
introduced a highly optimized, IO-aware algorithm that computes attention in a tiled fashion, thereby avoiding the 
materialization of large intermediate matrices. By tightly coordinating memory access with computation, FlashAttention 
achieves significant speedups and memory savings, enabling training with longer sequence lengths and larger batch sizes. 
This line of work illustrates how re-engineering core components of the transformer architecture \cite{vaswani2023attentionneed} 
can unlock substantial efficiency gains and expand the practical limits of large-scale model training.

\subsection{Operator Fusion in LLM Systems}
Operator fusion has also been widely adopted in industry-grade libraries. NVIDIA's TransformerEngine \cite{transformerengine} 
provides a suite of fused CUDA kernels tailored for transformer workloads \cite{vaswani2023attentionneed}, including GEMM 
combined with bias addition and activation functions, fused layer normalization, and optimized softmax variants. By collapsing 
multiple sequential operations into a single kernel, TransformerEngine eliminates redundant memory movement and reduces kernel 
launch overhead, thereby improving arithmetic intensity and throughput. These fused primitives are designed to exploit Tensor 
Cores and mixed-precision formats such as FP8 and BF16, enabling efficient training and inference of very large language models. 
TransformerEngine exemplifies how production systems integrate operator fusion into the transformer stack, complementing 
research efforts like Online Softmax and FlashAttention, and reinforcing the broader trend of re-engineering fundamental 
components for efficiency.

Together, these efforts highlight a broader trajectory: efficiency gains in LLM training often arise from collapsing 
boundaries between traditionally separate operations. In line with this trajectory, we introduce a kernel-level optimization 
that pushes forward the broader agenda of reducing memory movement and improving arithmetic intensity in LLM systems.

\section{Methodology}
\label{sec:methodology}

\subsection{Standard Output Projection and Loss Prediction}
In conventional LLM training, the output layer consists of two sequential operations:

\begin{enumerate}[label=\alph*)]

\item Projection into vocabulary logits

Given hidden states $H \in R^{B \times T \times d}$ and $W \in R^{V \times d}$, where $B$ is the batch size, 
$T$ is the sequence length, $d$ is the hidden dimension, and $V$ is the vocabulary size, the logits are computed as:

\begin{equation}
\label{eq:projection}
Z = H \cdot W^T
\end{equation}
where $\cdot$ denotes the matrix multiplication. $Z \in R^{B \times T \times V}$ is the resulting logits tensor.

\item Cross-entropy loss prediction

For target tokens $Y \in \{1, 2, ..., V\}^{B \times T}$, the loss is:

\begin{equation}
\label{eq:prediction}
\mathcal{L} = - \frac{1}{B T} \sum_{b=1}^{B} \sum_{t=1}^{T} 
\log \frac{\exp(Z_{b,t,Y_{b,t}})}{\sum_{v=1}^{V} \exp(Z_{b,t,v})}
\end{equation}

This formulation requires full materialization of $Z$, which scales as $O(B\cdot T\cdot V)$. For large 
vocabularies, this tensor dominates GPU memory and bandwidth usage, even though only one logit per position
$(Z_{b,t,Y_{b,t}})$ is ultimately consumed.

\end{enumerate}

\subsection{Proposed Approach}
Our key insight is that the cross-entropy loss can be computed directly from hidden states and target tokens
without explicitly forming the full logits tensor. Specifically:

\begin{equation}
\label{eq:direct_loss}
\mathcal{L} = - \frac{1}{B T} \sum_{b=1}^{B} \sum_{t=1}^{T} 
(H_{b,t} \cdot W^T_{Y_{b,t}} - \log \sum_{v=1}^{V} \exp(H_{b,t} \cdot W^T_v))
\end{equation}

\begin{itemize}
\item $H_{b,t} \cdot W^T_{Y_{b,t}}$ computes only the logit corresponding to the target token at position $(b,t)$
\item The dominator term is evaluated via a streaming reduction, avoiding storage of all intermediate logits
\end{itemize}

This fused formulation eliminates the need to materialize $Z$, reducing memory footprint from 
$O(B\cdot T\cdot V)$ to $O(B\cdot T)$, while maintaining the exact quivalence to the standard two-stage pipeline.

To ensure numerical stability while avoiding full logits materialization, our fused kernel employs a streaming 
variant of the softmax computation. Instead of storing all vocabulary logits, the algorithm maintains two running 
variables per position: the current maximum logit and an accumulator of exponentials relative to that maximum. 
As each vocabulary entry is processed, the maximum and accumulator are updated simultaneously, guaranteeing 
stability against overflow or underflow. The target logit corresponding to the ground-truth token is tracked 
separately for the numerator, while the denominator is reconstructed at the end as the product of the 
maximum's exponential and the accumulated sum. This design allows the loss to be computed exactly as 
in the standard cross-entropy formulation, but without ever instantiating the full $(B\cdot T\cdot V)$
logits tensor. By combining projection and loss computation in a single pass, the algorithm reduces memory footprint, 
alleviates bandwidth pressure, and preserves correctness under safe softmax semantics. Algorithm \ref{alg:fused_forward}
details the implementation of the fused kernel.

\begin{algorithm}[H]
\caption{Fused Output Projection and Cross-Entropy Prediction with Safe Softmax}
\label{alg:fused_forward}
\begin{algorithmic}[1]
\REQUIRE Hidden states $H \in \mathbb{R}^{B \times T \times d}$, 
         Output weights $W \in \mathbb{R}^{V \times d}$, 
         Target tokens $Y \in \{1,\dots,V\}^{B \times T}$
\ENSURE Cross-entropy loss $\mathcal{L}$

\STATE Initialize $\mathcal{L} \gets 0$
\FOR{$b = 1$ to $B$}
  \FOR{$t = 1$ to $T$}
    \STATE Initialize $m \gets -\infty$ \COMMENT{running maximum}
    \STATE Initialize $a \gets 0$ \COMMENT{accumulated sum}
    \STATE Initialize $z_{\text{target}} \gets 0$
    \FOR{$v = 1$ to $V$}
      \STATE Compute logit: $z \gets H_{b,t} \cdot W_v^\top$
      \IF{$z > m$}
        \STATE $a \gets a \cdot \exp(m - z) + 1$
        \STATE $m \gets z$
      \ELSE
        \STATE $a \gets a + \exp(z - m)$
      \ENDIF
      \IF{$v = Y_{b,t}$}
        \STATE $z_{\text{target}} \gets z$
      \ENDIF
    \ENDFOR
    \STATE Compute denominator: $s \gets \exp(m) \cdot a$
    \STATE Update loss: 
           $\mathcal{L} \gets \mathcal{L} - \log \frac{\exp(z_{\text{target}})}{s}$
  \ENDFOR
\ENDFOR
\STATE Normalize: $\mathcal{L} \gets \mathcal{L} / (B \cdot T)$
\RETURN $\mathcal{L}$
\end{algorithmic}
\end{algorithm}

For clarity of presentation, we detail the forward algorithm here, while the backward pass is described 
separately in Appendix \ref{app:backward_pass}.

\begin{figure}[htbp]
    \centering
    \includegraphics[width=0.9\linewidth]{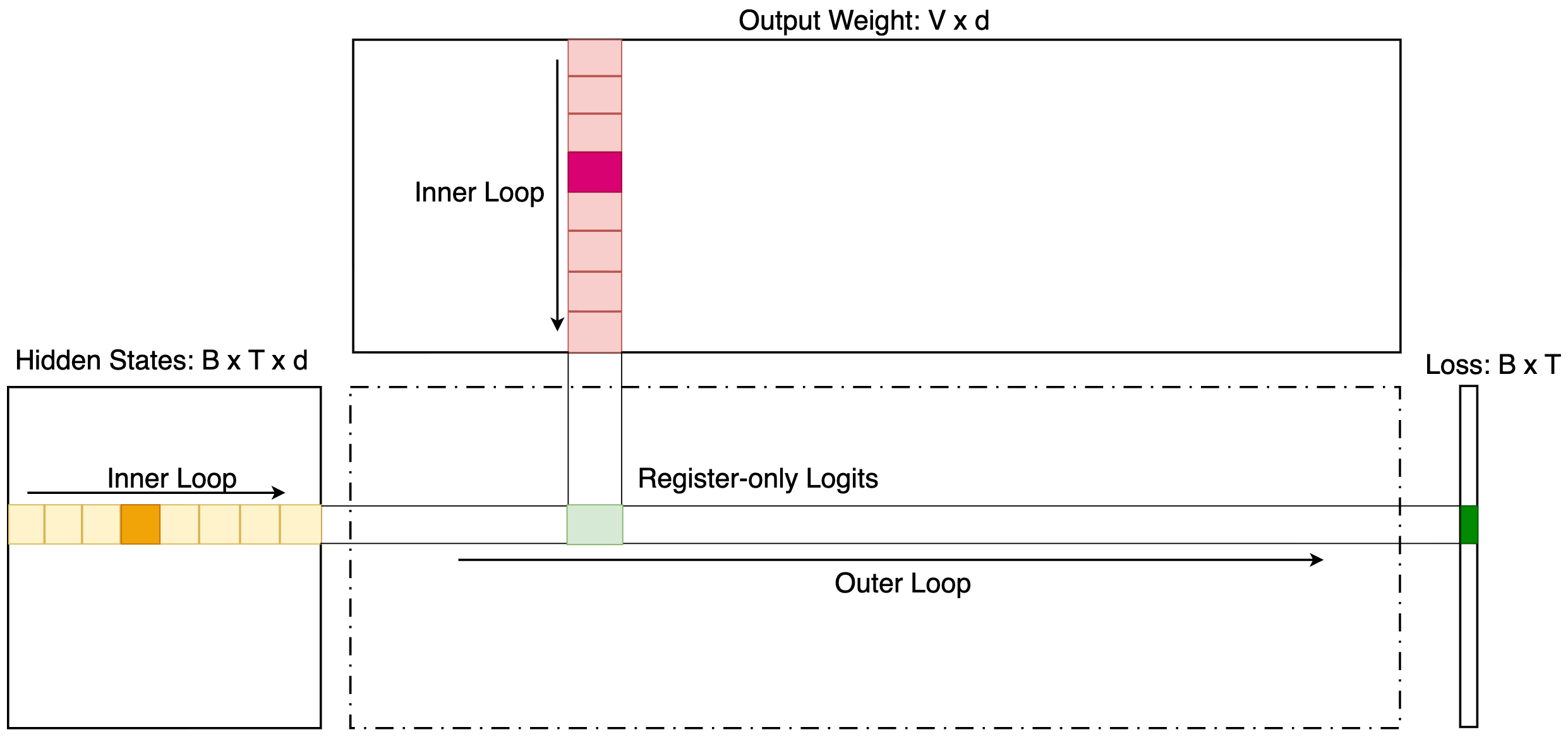}
    \caption{GPU-friendly forward pass with register-level logits and parallel accumulation updates}
    \label{fig:fwd}
\end{figure}

\begin{figure}[hptb]
    \centering
    \includegraphics[width=0.9\linewidth]{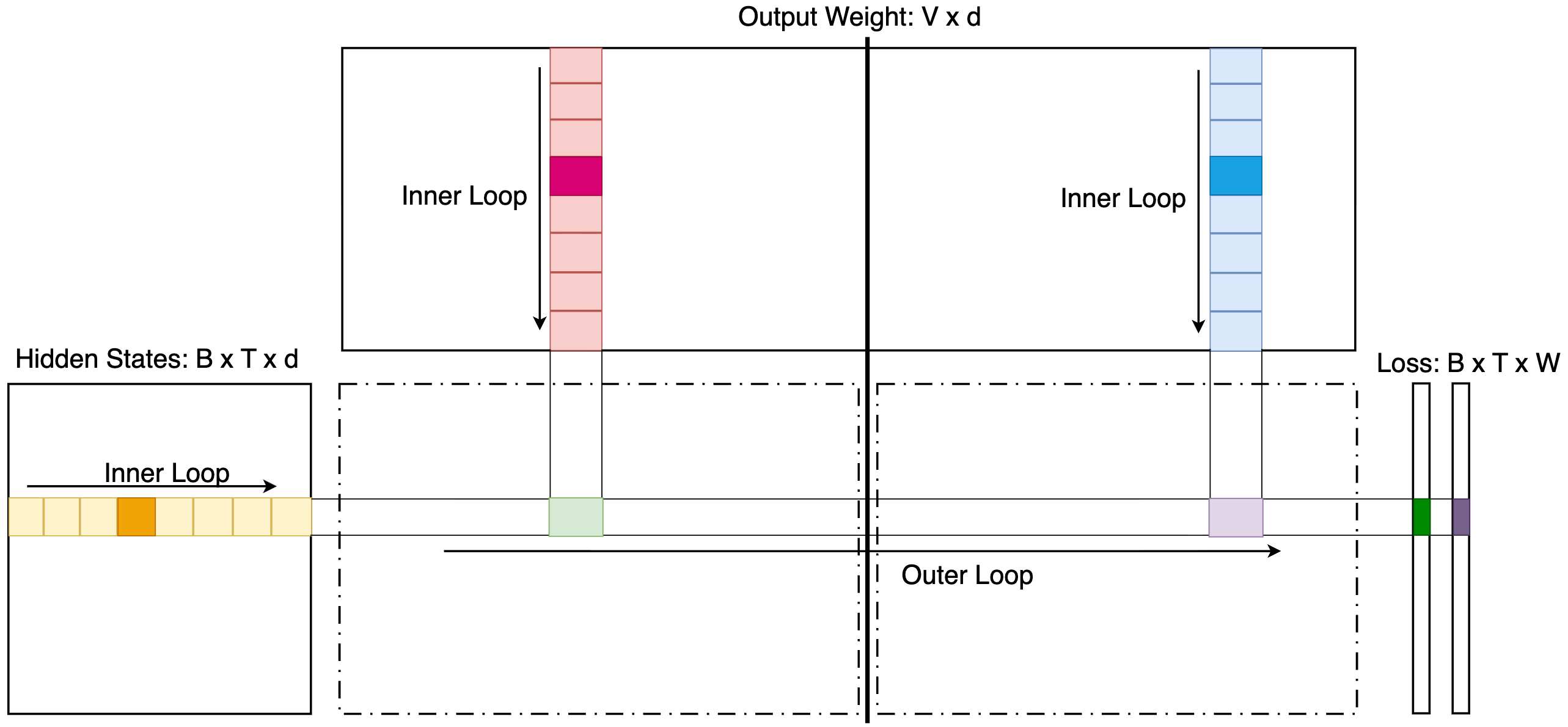}
    \caption{Window-based forward pass with parallel accumulation updates}
    \label{fig:parallel_fwd}
\end{figure}

To visually illustrate the projection-prediction fusion, Figure~\ref{fig:fwd} depicts the key computation steps in 
the forward pass. Yellow tiles represent slices of the hidden state tensor, while pink tiles denote segments of the output 
weight matrix assigned to each thread block. The diagram emphasizes the nested loop structure of the output layer: 
the inner loop performs dot products between hidden states and output weights, and the outer loop accumulates 
contributions across the vocabulary dimension. This tile-based execution enables parallel updates to the running maximum 
and normalization accumulator, two critical components of the numerically stable softmax. The design is well-suited 
to GPU architecture, leveraging warp-level parallelism and register-local reductions. The intermediate logits tensor 
is shown as a dashed box, indicating that it is never materialized in device memory but instead computed and consumed 
on-the-fly within GPU registers.

\subsubsection{Window-based Strategy}
In the vanilla design, each thread block performs accumulation across the vocabulary dimension in parallel for a given $(b, t)$ position.
While this approach is effective when the product $B \times T$ is large, it suffers from poor GPU occupancy when $B \times T$ is small - 
particularly in scenarios with a large vocabulary size $V$, where the number of active blocks becomes insufficient to saturate the device.

To mitigate this limitation, we introduce a tunable hyperparameter called the \emph{window size}, denoted by $W$, which partitions
the outer loop into multiple chunks. Each window is processed independently by a group of blocks, enabling finer-grained parallelism and better
utilization of GPU resources. Figure~\ref{fig:parallel_fwd} illustrates this window-based strategy for enhancing occupancy 
during the forward pass. The diagram shows two distinct windows, each representing a chunk of the outer loop split along the vocabulary axis. 
By assigning separate groups of thread blocks to process each window independently, this design increases parallel granularity and ensures high 
GPU utilization even when $B \times T$ is small. This approach allows concurrent accumulation across the vocabulary dimension within each window,
improving throughput without altering the computational semantics.

In order to obtain final prediction result, an additional epilogue operation is required to aggregate the partial outputs from all windows.

\subsubsection{Model Parallelism}

\begin{figure}[hptb]
    \centering
    \includegraphics[width=0.7\linewidth]{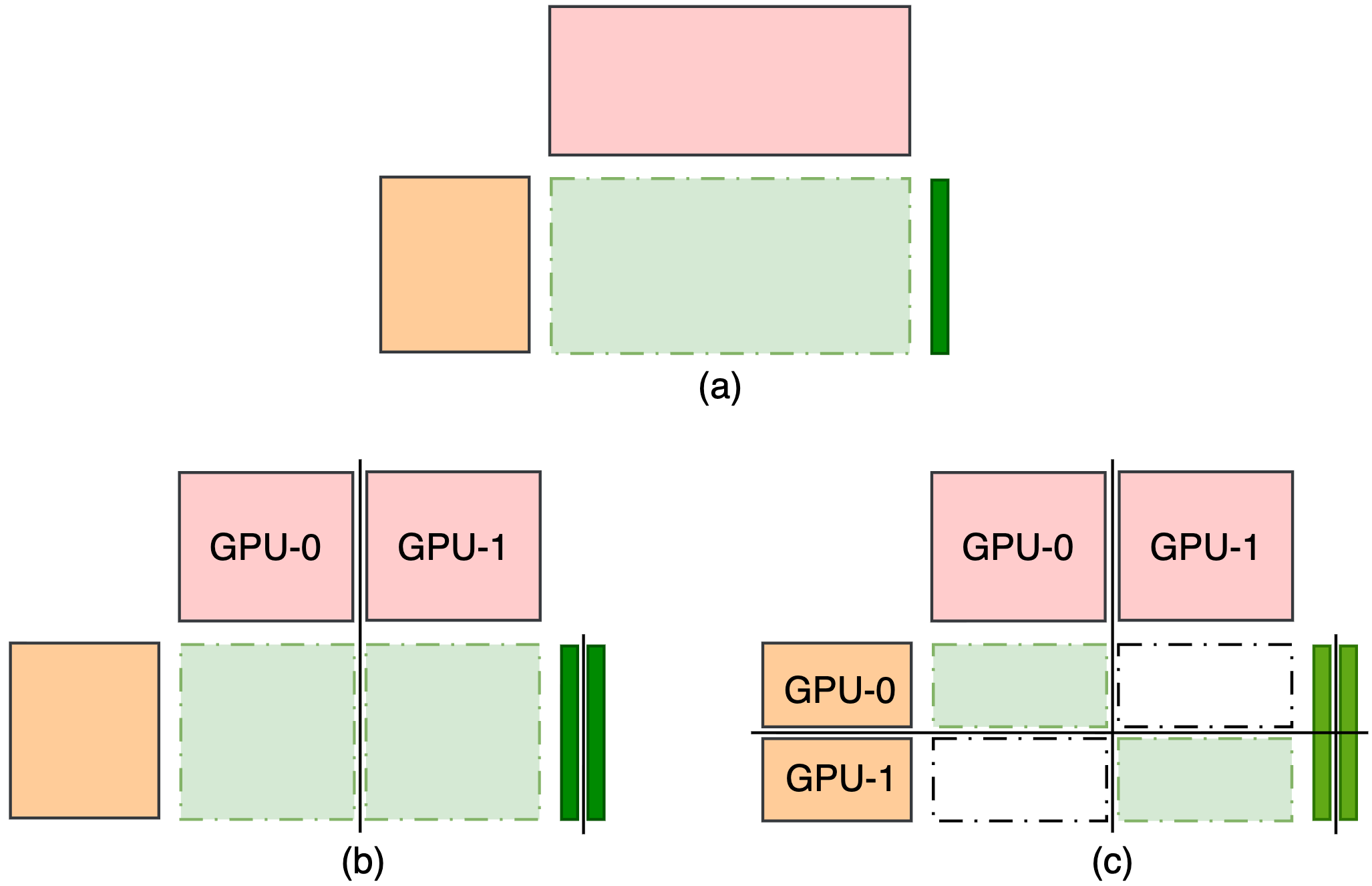}
    \caption{Model parallelism patterns}
    \label{fig:model_parallelism}
\end{figure}

The proposed method is readily extensible to support a variety of model parallelism strategies \cite{megatron}. Among the most commonly
adopted patterns in large-scale language model training are Data Parallelism (DP), Tensor Parallelism (TP), and Sequence Parallelism (SP), each
targeting different bottlenecks in memory and compute distribution. 

Figure~\ref{fig:model_parallelism} illustrates how our work integrates
with these parallelism schemes. In this figure, yellow boxes represent hidden states, pink boxes denote output weights, 
light green boxes indicate register-only logits, and dark green boxes correspond to the final loss predictions.

\begin{enumerate}

\item Data Parallelism, Figure~\ref{fig:model_parallelism}(a)

In DP, each GPU processes a distinct mini-batch of input data while maintaining a full copy of the model parameters, including hidden states 
and output weights. Gradients are computed independently on each device and synchronized across ranks during the backward pass. This strategy
is widely adopted due to its simplicity and scalability. Our fused kernels integrate seamlessly into this setup, requiring no changes to the DP
workflow.

\item Tensor Parallelism, Figure~\ref{fig:model_parallelism}(b)

TP partitions the output weight matrix along the vocabulary axis across GPU ranks. Each rank computes logits and partial predictions 
for its assigned vocabulary shard. Our fused kernel supports this configuration by restricting the outer loop to the local 
weight slice and accumulating partial results accordingly. To produce the final prediction, partial outputs must be aggregated across
all TP ranks.

\item Sequence Parallelism, Figure~\ref{fig:model_parallelism}(c)

SP further divides the hidden states along the sequence axis, typically layered on top of TP. Each GPU processes a subset of the 
sequence positions while retaining a shard of the output weights. Our design accommodates this setup by first gathering partial hidden states
and then convert the SP layout into a TP-compatible pattern. This ensures that all SP ranks produce consistent prediction results without
introducing semantic divergence.

\end{enumerate}

\section{Experiments}
\label{sec:experiments}

To evaluate the effectiveness of the proposed projection-prediction combination method, we conduct a series of experiements measuring both
runtime performance and memory consumption. Our goal is to quantify the benefits of eliminating the logits materialization and leveraging
register-local computation, particularly in regimes where vocabulary size and sequence length pose scalability challenges.

\subsection{Experimental Setup}

We implement these kernels with NVIDIA CUTEDSL \cite{cutedsl} and OpenAI Triton \cite{triton}, and incorporate them into the PyTorch-based 
training pipeline. To verify its effectiveness, we compare it against the canonical two-stage implementation that separately computes logits
and applies cross-entropy loss \cite{10888877}. All experiments are conducted on NVIDIA GB200 GPU with 186GB memory, 
using mixed-precision training (BF16). We benchmark across multiple batch sizes, sequence lengths, and vocabulary sizes to capture 
a range of realistic training scenarios.

Table~\ref{tab:experimental_setup} summarizes the experimental settings.

\begin{table}[h]
\centering
\caption{Experimental setup}
\label{tab:experimental_setup}
\begin{tabular}{@{}lccc@{}}
\toprule
Aspect & Config \\
\midrule
Data Type & BF16 \\
GPU & NVIDIA GB200 \\
PyTorch  & 2.9.0 \\
OpenAI Triton & 3.4.0 \\
NVIDIA CUTEDSL & 4.2.1 \\
Hidden Dimension, $d$ & 4096 \\
Range of $B \times T$ & 1024,4096,8192,16384,32768 \\
Range of $V$ & 32768,65536,131072,262144 \\
Model Parallelism & Data Parallelism \\
\bottomrule
\end{tabular}
\end{table}

Since the input datatype is BF16 \cite{nvidiabf16}, intermediate logits must be upcast to FP32 \cite{IEEE754} during computation 
to safeguard against numerical underflow and overflow. In the canonical two-stage approach, this upcasting occurs within the GEMM operation,
where the full logits tensor is materialized in device memory. In contrast, our method inherently avoids this overhead by performing 
the upcast within register-local computation.

\subsection{Results}

\begin{table}[h]
  \centering
  \caption{Latency (ms) and memory usage (MB) comparison between canonical and proposed methods across varying 
           $B \times T$ and vocabulary sizes $V$ on Data-Parallelism. The lower, the better.}
  \label{tab:dp_results}
  \begin{tabular}{c c cc cc}
  \toprule
  \multirow{2}{*}{$B \times T$} & \multirow{2}{*}{$V$} & \multicolumn{2}{c}{Latency (ms)} & \multicolumn{2}{c}{Memory (MB)} \\
  \cmidrule(lr){3-4} \cmidrule(lr){5-6}
  & & Canonical & Proposed & Canonical & Proposed \\
  \midrule
  \multirow{4}{*}{1024}   & 32,768   & 0.73 & \bfseries 0.69 & 1064  & \bfseries 280 \\
                          & 65,536   & 1.40 & \bfseries 0.97 & 2088  & \bfseries 536 \\
                          & 131,072  & 2.76 & \bfseries 1.27 & 4136  & \bfseries 1048 \\
                          & 262,144  & 5.46 & \bfseries 1.83 & 8232  & \bfseries 2072 \\
  \midrule
  \multirow{4}{*}{4096}   & 32,768   & 1.66 & \bfseries 1.16 & 1904  & \bfseries 304 \\
                          & 65,536   & 3.34 & \bfseries 1.69 & 3696  & \bfseries 561 \\
                          & 131,072  & 6.78 & \bfseries 2.86 & 7280  & \bfseries 1073 \\
                          & 262,144  & 13.70 & \bfseries 5.22 & 14448 & \bfseries 2099 \\
  \midrule
  \multirow{4}{*}{8192}   & 32,768   & 2.93 & \bfseries 1.70 & 3024  & \bfseries 337 \\
                          & 65,536   & 6.08 & \bfseries 2.90 & 5840  & \bfseries 593 \\
                          & 131,072  & 13.25 & \bfseries 5.58 & 11472 & \bfseries 1107 \\
                          & 262,144  & 27.77 & \bfseries 10.79 & 22736 & \bfseries 2133 \\
  \midrule
  \multirow{4}{*}{16384}  & 32,768   & 5.62 & \bfseries 3.43 & 5264  & \bfseries 401 \\
                          & 65,536   & 12.41 & \bfseries 6.61 & 10128 & \bfseries 659 \\
                          & 131,072  & 26.67 & \bfseries 13.20 & 19856 & \bfseries 1173 \\
                          & 262,144  & 50.15 & \bfseries 25.43 & 39312 & \bfseries 2203 \\
  \midrule
  \multirow{4}{*}{32768}  & 32,768   & 11.42 & \bfseries 7.00 & 9744  & \bfseries 531 \\
                          & 65,536   & 25.64 & \bfseries 13.80 & 18704 & \bfseries 790 \\
                          & 131,072  & 50.24 & \bfseries 27.06 & 36624 & \bfseries 1307 \\
                          & 262,144  & 96.52 & \bfseries 53.61 & 72464 & \bfseries 2342 \\
  \bottomrule
  \end{tabular}
\end{table}

\begin{figure}[htbp]
  \centering
  \includegraphics[width=0.9\linewidth]{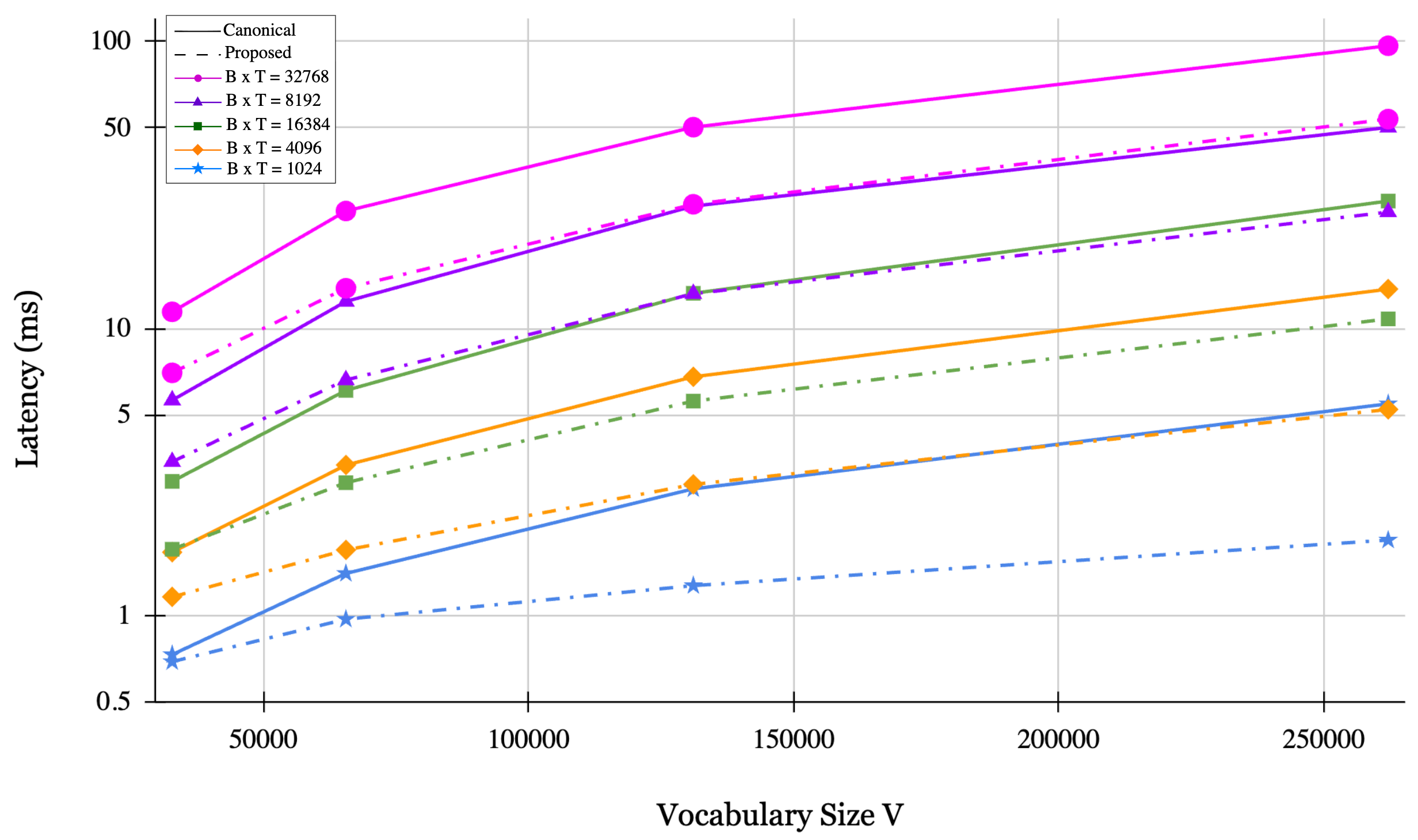}
  \caption{Latency Comparison between Canonical and Proposed Methods. The lower, the better.}
  \label{fig:dp_latency}
\end{figure}

\begin{figure}[htbp]
  \centering
  \includegraphics[width=0.9\linewidth]{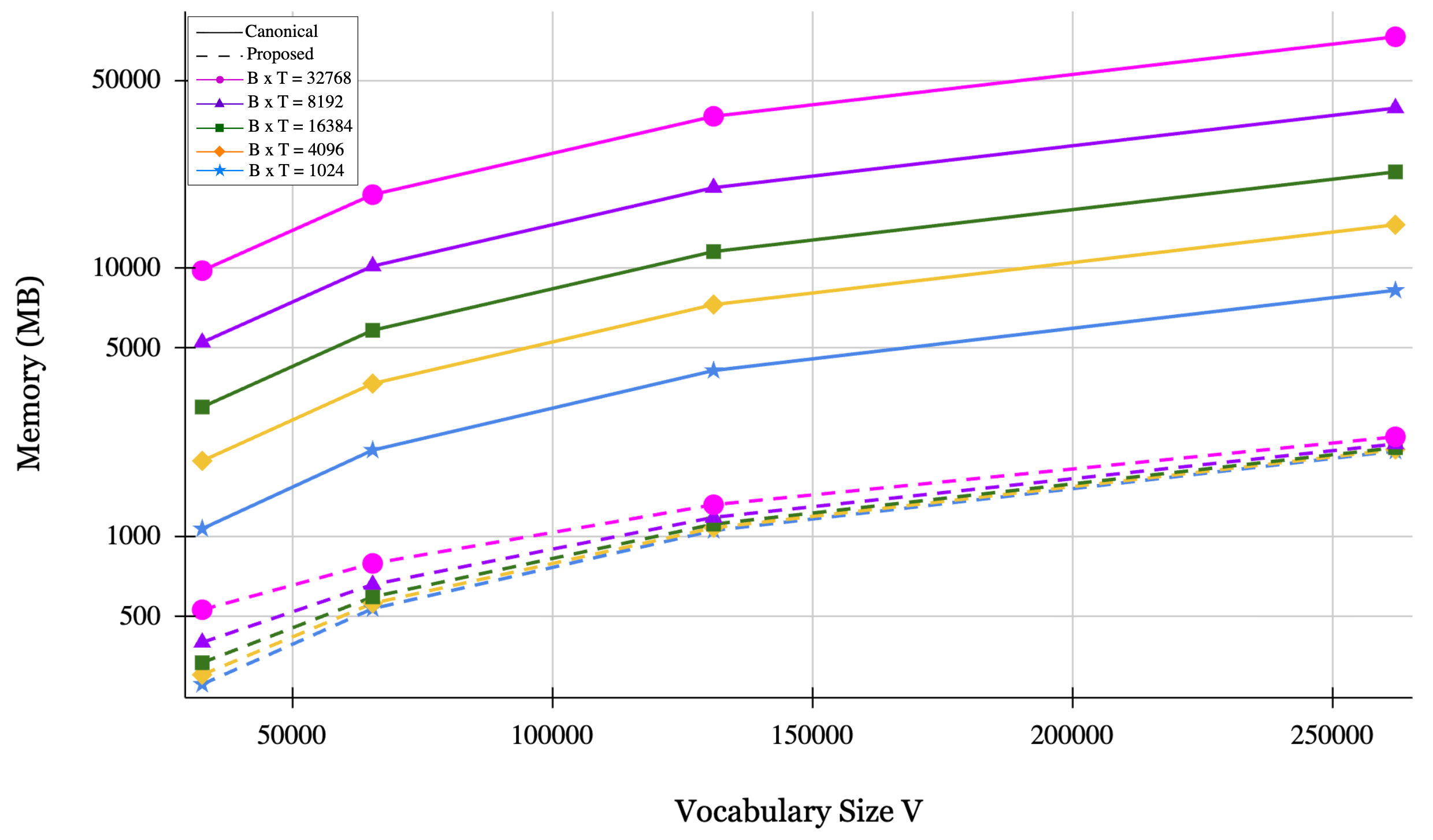}
  \caption{Memory Comparison between Canonical and Proposed Methods. The lower, the better.}
  \label{fig:dp_memory}
\end{figure}

Table~\ref{tab:dp_results} presents a comprehensive comparison between the canonical and proposed method across varying vocabulary
size $V$ and batch-sequence products $B \times T$ under Data Parallelism. Figure~\ref{fig:dp_latency} visualizes the latency trends,
while Figure~\ref{fig:dp_memory} illustrates memory usage. Together, these results demonstrate that our proposed method consistently
outperforms the canonical two-stage implementation in both runtime and memory efficiency.

Latency, measured in milliseconds, increases with vocabulary size due to the growing cost of output projection and loss computation. 
This trend holds across all $B \times T$ configurations. However, our kernel significantly mitigates this growth. For instance, at
$B \times T =$ 32,768 and $V =$ 262,144, the canonical method incurs a latency of 96.42 ms, whereas our method reduces it to 53.61 ms - 
yielding a \textbf{44.5\%} improvement. Figure~\ref{fig:dp_latency} illustrates this pattern, showing that the proposed method maintains
a consistently lower latency curve across all vocabulary sizes.

Memory usage, gauged in megabytes, also scales linearly with vocabulary size in the canonical implementation due to full logits tensor
materialization. In contrast, our method avoids this overhead by computing logits in a register-local fashion. 
As shown in Table~\ref{tab:dp_results}, the memory savings are substantial. At $B \times T =$ 32,768 and $V =$ 262,144, the canonical 
method consumes 72.5 GB, while our method requires only 2.3 GB - a reduction of over \textbf{96.8\%}. Figure~\ref{fig:dp_memory} highlights
this efficiency, with the proposed method exhibiting a much flatter memory growth curve.

Across all tested configurations, the proposed method delivers consistent improvements in both latency and memory footprint. 
These gains become increasingly pronounced at larger vocabulary sizes and longer sequence lengths, making our approach particularly 
well-suited for scaling LLM training in memory-constrained or latency-sensitive environments.

\section{Discussion}
\label{sec:discussion}

The experimental results confirm that our fused projection-prediction kernel consistently outperforms the canonical two-stage implementation 
in both latency and memory usage across diverse vocabulary sizes and sequence configurations. These improvements are most pronounced in large-scale 
settings, where the cost of output projection and logits materialization emerges as a dominant bottleneck.

The performance improvements arise from collapsing the boundary between the projection and loss stages. By fusing these operations, 
the proposed method achieves finer-grained GPU resource utilization and better overlap between CUDA Core and Tensor Core execution. 
This design reduces kernel launch overhead, increases arithmetic intensity, and ultimately delivers higher throughput. Furthermore, 
by eliminating the need to materialize the full logits tensor in device memory, our method alleviates bandwidth pressure and avoids costly 
I/O associated with reading and writing large activation tensors.

In the canonical approach, a large intermediate logits tensor must be stored, scaling with $B \times T \times V$ and imposing significant 
memory and bandwidth demands. Our method sidesteps this by computing logits on-the-fly within register-local scopes, enabling execution 
with a lower memory footprint and reduced data movement. This design aligns naturally with the architectural strengths of modern GPUs, 
which favor compute-bound workloads with minimized memory traffic. The result is a streamlined execution pipeline that scales gracefully 
with vocabulary size.

While the fused kernel offers clear advantages, it introduces additional complexity in kernel design and debugging. Numerical stability 
must be carefully managed, particularly under mixed-precision regimes (e.g., BF16 inputs with FP32 accumulation), where rounding 
behavior can affect convergence. Moreover, although the NVIDIA CUDA toolkit provides the flexibility required for fine-grained register 
control and efficient tensor core scheduling, the fusion strategy may be less portable to programming environments - such as OpenAI Triton - 
that abstract away low-level resource management. These are not hardware limitations, but software constraints imposed by 
higher-level compilation frameworks.

Although our experiments focus on the standard cross-entropy loss, the proposed combining method is not restricted to this objective. 
By collapsing the boundary between projection and prediction, the fused design generalizes naturally to more complex output layers, 
including multi-task heads, structured prediction objectives, and loss variants such as label smoothing or sampled softmax. In all 
these cases, the principle of avoiding logits materialization and leveraging register-local computation applies, enabling efficient 
scaling without fundamental redesign. This extensibility underscores the broader applicability of our approach beyond 
standard training setups.

Future work will extend the fusion strategy to support additional loss functions such as label smoothing and sampled softmax. 
Our current implementation already supports tensor parallelism and sequence parallelism; ongoing experiments will further 
validate its effectiveness and scalability in multi-GPU and multi-node environments. Another promising direction is 
automated kernel generation using compiler-based approaches, specifically within the NVIDIA CUDA ecosystem. Rather than 
targeting cross-architecture portability, we aim to deepen optimization within NVIDIA's GPU stack, reducing manual 
kernel engineering while preserving fine-grained control over register usage and tensor core scheduling.

\section{Conclusion}
\label{sec:conclusion}

Training large language models at scale is increasingly constrained by the cost of output projection and loss computation, 
particularly as vocabulary sizes and sequence lengths grow. The canonical two-stage approach suffers from high latency and 
excessive memory usage due to the materialization of large logits tensors, creating a bottleneck in both performance and scalability.

To address this challenge, we proposed a fused projection-prediction kernel that collapses the boundary between projection and 
loss computation. By computing logits on-the-fly within register-local scopes, our method eliminates the need to store intermediate 
tensors, reduces memory bandwidth pressure, and enables finer-grained utilization of GPU resources. This design achieves better overlap 
between CUDA Core and Tensor Core execution, improving arithmetic intensity and throughput.

Experimental results demonstrate that the proposed method consistently outperforms the canonical implementation across a wide range 
of vocabulary sizes and batch-sequence configurations. At large scales, the improvements are particularly striking: latency reductions 
of over 40\% and memory savings exceeding 95\% were observed. These gains highlight the effectiveness of our approach in mitigating 
the dominant bottlenecks of LLM training.

Beyond cross-entropy loss, the fused design generalizes naturally to more complex objectives, including multi-task heads, 
structured prediction, and loss variants such as label smoothing or sampled softmax. This extensibility underscores the broader 
applicability of our method across diverse training regimes.

Looking ahead, future work will focus on further validating the integration of our fused kernel with tensor and sequence 
parallelism to assess scalability in multi-GPU and multi-node environments. We also see compiler-assisted kernel generation - 
within the NVIDIA CUDA ecosystem - as a promising direction to automate and extend the fusion strategy across diverse model architectures 
and training regimes. Rather than pursuing cross-platform generalization, our goal is to deepen optimization within NVIDIA's software stack, 
reducing engineering overhead while preserving fine-grained control. Ultimately, our approach contributes to more efficient 
and scalable training pipelines, supporting the continued evolution of large language models under increasingly demanding 
computational constraints.

\section*{Acknowledgments}

We would like to thank NVIDIA DevTech team for valuable comments and suggestions.

\bibliography{references}

\appendix
\section{Appendix}
\label{app:appendix}

\subsection{Backward Pass}
\label{app:backward_pass}
For each position $(b,t)$, let $z_v = H_{b, t} \cdot W_v^T$ and $P_v = \frac{exp(z_v)}{\sum_{v=1}^{V} exp(z_v)}$. 
The cross-entropy gradient w.r.t. logits is $\frac {\partial L} {\partial z_v} = P_v - 1[v == Y_{b,t}]$. 
In the fused setup, we obtain $P_v$ via the same streaming, numerically stable softmax used in the forward pass: 
maintain the running maximum $m$ and accumulator $a$, then compute $P_v = \exp(z_v - m) / a$ on the fly for each $v$.
Gradients propagate without materializing the logits tensor, 

the gradient w.r.t. hidden states is 

\begin{equation}
\label{eq:backward_hidden}
\frac {\partial L} {\partial H_{b,t}} = \sum_{v=1}^{V} \frac {\partial L} {\partial z_v} \cdot W_v 
= \sum_{v=1}^{V} (P_v - 1[v == Y_{b,t}]) \cdot W_v
\end{equation}

and the gradient w.r.t. output weights accumulates as

\begin{equation}
\label{eq:backward_weights}
\frac {\partial L} {\partial W_v} = \sum_{b=1}^{B} \sum_{t=1}^{T} \frac {\partial L} {\partial z_v} \cdot H_{b,t}
= \sum_{b=1}^{B} \sum_{t=1}^{T} (P_v - 1[v == Y_{b,t}]) \cdot H_{b,t}
\end{equation}

Operationally, the kernel streams over $v$, to re-compute forward logit $z_v$ and then compute $P_v$ stably using $(m, a)$, 
updates $\partial L / \partial H_{b,t}$ 
by accumulating $(P_v - 1[v == Y_{b,t}])W_v$, and atomically accumulates $\partial L / \partial W_v$ with
$(P_v - 1[v == Y_{b,t}])H_{b,t}$. This preserves exact gradients while avoiding storage of $Z \in R^{B\cdot T\cdot V}$.

Algorithm \ref{alg:fused_backward} details the implementation of the backward pass.
\begin{algorithm}[H]
    \caption{Backward Pass with Logit Recompute and Safe Softmax}
    \label{alg:fused_backward}
    \begin{algorithmic}[1]
    \REQUIRE Hidden states $H$, Output weights $W$, Targets $Y$,
             Safe-softmax stats $(m,a)$ from forward,
             Upstream gradient $\Gamma$ (e.g., $\Gamma = 1/(B \cdot T)$ for mean reduction)
    \ENSURE Gradients $\partial H$, $\partial W$
    
    \STATE Initialize $\partial H \gets 0$, $\partial W \gets 0$
    \FOR{$b = 1$ to $B$}
      \FOR{$t = 1$ to $T$}
        \STATE $\partial H_{b,t} \gets 0$
        \FOR{$v = 1$ to $V$}
          \STATE $z \gets H_{b,t} \cdot W_v^\top$
          \STATE $p_v \gets \exp(z - m) / a$
          \STATE $g_v \gets \Gamma_{b,t} \cdot \left( p_v - \mathbf{1}[v = Y_{b,t}] \right)$
          \STATE $\partial H_{b,t} \gets \partial H_{b,t} + g_v \cdot W_v$
          \STATE $\partial W_v \gets \partial W_v + g_v \cdot H_{b,t}$
        \ENDFOR
      \ENDFOR
    \ENDFOR
    \RETURN $\partial H, \partial W$
    \end{algorithmic}
\end{algorithm}

Furthermore, there exists a more radical strategy to achieve the same objective. In this variant, the training 
pipeline is restructured to perform \textbf{partial gradient accumulation} for both hidden states and output weights
directly within the forward pass. The subsequent backward pass then applies the upstream gradient to rescale these
partial results, producing the final gradients. By design, this approach eliminates the need to recompute the logits
during backpropagation and thereby avoids redundant matrix multiplications, substantially reducing computational overhead.
\textbf{It is important to note, however, that this method is only applicable when the upstream gradient is a scalar - 
namely, when the loss reduction mode is set to either mean or sum.}

Algorithm \ref{alg:fused_forward_partial_gradient} presents the forward pass with integrated partial gradient 
accumulation, while Algorithm \ref{alg:backward_scaling} illustrates the subsequent backward scaling step that 
incorporates the upstream gradient.

\begin{algorithm}[H]
    \caption{Forward with Safe Softmax + Partial Gradient Accumulation}
    \label{alg:fused_forward_partial_gradient}
    \begin{algorithmic}[1]
    \REQUIRE $H \in \mathbb{R}^{B \times T \times d}$, $W \in \mathbb{R}^{V \times d}$, $Y \in \{1,\dots,V\}^{B \times T}$
    \ENSURE Loss $\mathcal{L}$, partial grads $\partial' H$, $\partial' W$; cached stats $(m,a)$
    
    \STATE $\mathcal{L} \gets 0$, $\partial' H \gets 0$, $\partial' W \gets 0$
    \FOR{$b=1$ to $B$}
      \FOR{$t=1$ to $T$}
        \STATE $m \gets -\infty$, $a \gets 0$, $z_{\text{target}} \gets 0$
        \FOR{$v=1$ to $V$}
          \STATE $z \gets H_{b,t} \cdot W_v^\top$
          \IF{$z > m$}
            \STATE $a \gets a \cdot \exp(m - z) + 1$; $m \gets z$
          \ELSE
            \STATE $a \gets a + \exp(z - m)$
          \ENDIF
          \IF{$v = Y_{b,t}$}
            \STATE $z_{\text{target}} \gets z$
          \ENDIF
        \ENDFOR
        \STATE $s \gets \exp(m) \cdot a$
        \STATE $\mathcal{L} \gets \mathcal{L} - \log \left( \exp(z_{\text{target}}) / s \right)$
    
        \STATE \textbf{Partial gradient accumulation (unscaled by upstream)}
        \STATE $\partial' H_{b,t} \gets 0$
        \FOR{$v=1$ to $V$}
          \STATE $z \gets H_{b,t} \cdot W_v^\top$ \COMMENT{recompute or cache per-tile}
          \STATE $p_v \gets \exp(z - m)/a$
          \STATE $g'_v \gets p_v - \mathbf{1}[v = Y_{b,t}]$
          \STATE $\partial' H_{b,t} \gets \partial' H_{b,t} + g'_v \cdot W_v$
          \STATE $\partial' W_v \gets \partial' W_v + g'_v \cdot H_{b,t}$
        \ENDFOR
    
        \STATE Cache $(m,a)$ (optional if recomputing in backward is cheap)
      \ENDFOR
    \ENDFOR
    \RETURN $\mathcal{L}, \partial' H, \partial' W, (m,a)$
    \end{algorithmic}
\end{algorithm}

\begin{algorithm}[H]
    \caption{Backward Scaling with Upstream Gradient}
    \label{alg:backward_scaling}
    \begin{algorithmic}[1]
    \REQUIRE Partial gradients $\partial' H$, $\partial' W$; upstream gradient $\Gamma$
    \ENSURE Final gradients $\partial H$, $\partial W$
    
    \IF{ $\Gamma$ is scalar }
      \STATE $\partial H \gets \Gamma \cdot \partial' H$
      \STATE $\partial W \gets \Gamma \cdot \partial' W$
    \ELSE
      \STATE Initialize $\partial H \gets 0$, $\partial W \gets 0$
      \FOR{$b=1$ to $B$}
        \FOR{$t=1$ to $T$}
          \STATE $\alpha \gets \Gamma_{b,t}$
          \STATE $\partial H_{b,t} \gets \alpha \cdot \partial' H_{b,t}$
          \STATE \textbf{Note:} For weights, either store per-position contributions $\partial' W^{(b,t)}$ in forward,
          \STATE \quad then compute $\partial W \gets \sum_{b,t} \alpha \cdot \partial' W^{(b,t)}$ here.
        \ENDFOR
      \ENDFOR
    \ENDIF
    
    \RETURN $\partial H, \partial W$
    \end{algorithmic}
\end{algorithm}

\end{document}